\documentclass[twocolumn,fleqn,natbib]{svjour2}
\smartqed  % flush right qed marks, e.g. at end of proof
\usepackage{graphicx}
\usepackage{amsmath} % assumes amsmath package installed
\usepackage{amssymb}  % assumes amsmath package installed
\usepackage{xcolor}
\usepackage{tikz}
\journalname{Biological Cybernetics}
\begin{document}

\title{Prediction-for-CompAction: Navigation in social environments using generalized cognitive maps\thanks{This work has been supported by the former Spanish Ministry of Science and Innovation (project FIS2010-20054).
}
}
%\subtitle{\red{Generalized cognitive maps for social navigation in cooperative and non-cooperative environments}}


\author{Jose A. Villacorta-Atienza \and 
        Carlos Calvo \and 
        Valeri A. Makarov
}


\institute{J. A. Villacorta-Atienza \and C. Calvo \and V. A. Makarov* \at
Dept. of Applied Mathematics, Universidad Complutense de Madrid, Avda Complutense s/n, Madrid 28040, Spain \\
V. A. Makarov is also with the 
Instituto de Matem\'atica Interdiciplinar, 
Universidad Complutense de Madrid.\\
*\email{vmakarov@mat.ucm.es}           %  \\
% \emph{Present address:} of F. Author  %  if needed
% \and
% S. Author \at
% second address
}

%\date{Received: \today / Revised: date}
\date{Received: April 21, 2014 / Accepted: January 27 2015 / Published online: February 13, 2015}

\maketitle

\begin{abstract}

The ultimate navigation efficiency of mobile robots in human environments will depend on how we will appraise them: merely as impersonal machines or as human-like agents.  A functional machine should not expect help during its ordinary tasks, such as navigation in a crowd. In the same situation a humanoid agent may elicit cooperation and take advantage of recursive cognition, i.e. agent's actions will depend on decisions made by humans that in turn depend on the decisions made by the agent. To tackle this high-level cognitive skill we propose a neural network architecture implementing Prediction-for-CompAction paradigm. The network predicts possible human-agent interactions and compacts (eliminates) the time dimension of the given dynamic situation. The emerging static cognitive map can be readily used for planning actions. We provide numerical evidence that in many situations, including navigation in cluttered pedestrian flows, a humanoid agent can choose paths to targets significantly shorter than a cognitive robot treated as a functional machine. Moreover, the navigation safety, i.e. chances to avoid accidental collisions, increases under cooperation. Remarkably, these benefits yield no additional load to the mean society effort. Thus, the proposed navigation strategy is socially acceptable and the humanoid agent can behave as ``one of us". 
 
\end{abstract}

\section{Introduction}

One of the priorities of modern bio-inspired robotics and cognitive neuroscience is to make robots human-friendly, providing tolerable or even comfortable coexistence with people \citep{REF1}. Then cognitive abilities of the robots must be adequately designed according to their role in the human society \citep{Anthro}.  In this context, simulation of the robot navigation in human environments, e.g. collision avoidance in human crowds,  provides a testbed for examining different scenarios of the human-robot interaction.  
%For instance, according to anthropomorphism a human-like robot is expected to behave as a human, whereas  a functional robot can exhibit limited cognitive properties \citep{Hegel}.  

Heuristic algorithms enlightened the pivotal role of cooperation, but they did not tackle the problem of how cognition appears. A neural-network approach simulating cognition and cooperation as an emerging property (i.e. not rule-imposed) has been significantly less explored. Existing neural network models are mostly devoted to imitation of specific human actions and comprehension of human intentions \citep{REF19}. Architectures relying on a bottom-up approach to robot navigation still show limited cognitive abilities \citep{CRUSE2013}. Besides, existing neural networks mainly deal with non cooperative navigation (coexistence) \citep{BC2010,FRIZ}. 

From a more abstract point of view, Theory of Mind (ToM), i.e. a capacity of inferring the internal state and thoughts of a partner \citep{ToM}, provides a broader context for the human-robot cooperation. It has been argued that a ToM  robot will provoke more social reactions \citep{Appel} and will be capable of learning from an observer in the same way as human infants do \citep{Scassellati}. Using a simplistic neural network and evolutionary learning algorithms \cite{Kim} have shown that robots can represent other's self and learn from an observer. A recursive ToM has been used for algorithmic modeling of the collision-avoidance \citep{Takano}. It has been shown that the algorithm performance depends on the level of recursion.

The cooperation of ToM agents considering others as also having ToM leads to the emergence of \emph{recursive cognition} \citep{Recurs}: an action of an agent will depend on decisions made by other agents that in turn depend on the decisions made by the agent and so on. Therefore, a neural network exhibiting cooperative properties should encapsulate the dynamics of decision-making processes of similar or higher complexity according to recursion during cognition. It is a challenging problem barely explored in the literature. 

In this paper we study how the problem of social navigation, i.e. the movement of an agent through a space structured by the activity of humans, can be tackled by artificial cognition mechanisms named \textit{compact cognitive maps} \citep{BC2010}, which generalize basic cognitive processes existing in the mammal brain. We develop the work by considering two basic functional scenarios of the human-robot interaction present during social navigation: coexistence, when the agent does not receive any collaboration from humans during its movement, and cooperation, where both robot and human change their routes simultaneously to avoid collisions. We demonstrate that artificial cognition via compact cognitive maps naturally supports recursive cognition. It allows the cognitive agent to perceive the same social situation differently depending on if cooperation or coexistence is assumed, which leads to qualitatively different behaviors both in structured and unstructured crowds. We conclude that, in general, cooperation offers significant benefits. Remarkably, these benefits assume no additional load to the society effort, i.e. under cooperative navigation the agent does not destabilize the society and can behave as ``one of us". It is also remarkable however that coexistence (no cooperation from humans) is preferable in certain situations. Thus, in order to behave as a human, an artificial cognitive agent should be capable of switching between cooperative and coexistence strategies according to the context.

\subsection{Prediction-for CompAction paradigm}

In the last decades diverse experimental findings provided insight into the neural mechanisms of cognition involved in interaction with \textit{static} scenarios. It has been shown that animals for navigation in space use abstract representations of the environment named cognitive maps \citep{REF4,COGMAPS2013}. The cognitive maps act as a GPS, containing critical information for understanding the perceived space, as subject location (place cells),  objects obstructing free ways (boundary cells) and space metric (grid cells), but also about how to move through such space \citep{StaticGPS}. This concept has been successfully applied to robotics for supporting cognitive navigation but mostly limited to static environments \citep{StCogMaps, Biomimetic_review_2}.

However concerning cognition in time-changing scenarios, growing experimental evidence suggests that the neural structures generating cognitive maps (e.g. the hippocampus and medial prefrontal cortex) also participate in representation of dynamic situations. For instance, place cells coding the subject's position in space also encode speed, turning angle, and direction of moving objects \citep{Japoneses}. At larger scale, chemical damage of the rat hippocampus impairs avoidance of moving obstacles, with no effect on other critical abilities \citep{Checos}. Thus, in mammals, cognition of dynamic situations is built over cognitive maps and involves global network activity of entire brain areas.

Following experimental insight adaptive artificial cognitive maps partially solve the problem of internal representation of dynamic environments. However, their strai\-ghtforward application may lead to explosive growth of the required calculation power [see e.g. \citep{IEEE2013} and references therein]. 

In order to generalize the cognitive maps to dynamic environments, recently we have proposed an alternative approach, called Prediction-for-CompAction (PfCA) \citep{BC2010}. PfCA postulates that when we process a \emph{dynamic} situation our brain does not explicitly code the time dimension, but it extracts from the spatiotemporal information those critical events required for dealing with such situation and projects them to a \emph{static} (spatial) map named \textit{compact cognitive map}. 

In the context of navigation, these critical events will be the possible collisions with obstacles, so according to PfCA our brain evaluates where such collisions could occur and transform them into \textit{effective obstacles}, which are then structured as a compact cognitive map. It should be remarked that 1) in the ordinary cognitive maps the obstacles are real since the scenarios are static, and 2) in these static environments the compact cognitive maps are naturally reduced to the cognitive maps \citep{BC2010}. Therefore the compact cognitive maps would have the same properties than cognitive maps but for dynamic situations, so they would act as a dynamic GPS, providing us with   the information required to understand a time-changing situation but also to interact with it (e.g. providing navigation trajectories).

\subsection{Two complementary concepts for navigation in social environments}

\subsubsection{Coexistence}

In our previous works \citep{BC2010,INTECH2011} we have proposed and developed cognitive navigation based on PfCA in deterministic dynamic environments, whose evolution only depends on the dynamical initial conditions of their elements, but can not be altered by the agent's behaviour. Thus in these situations any recursive interaction between the agent and its environment, as cooperation or competition, is impossible. We have shown that, in these environments, the compact cognitive maps allow versatile navigation of robots, and that they serve as a natural substrate for crucial cognitive abilities as memory and learning \citep{IEEE2013}.

The essential concepts and results of these previous works are revisited in the context of robot navigation in social scenarios in the section \ref{CCM}. In these circumstances a robot regarded by humans merely as a functional machine should not expect any cooperation from people during its ordinary tasks, so it should move among people in coexistence, by considering its environment as dynamically deterministic. Most of the contemporary robot navigation strategies implicitly implement this non cooperative, or coexistence, relationship between the robot and its environment, assuming that it must ``make all the work'' to navigate safely, avoiding collisions  \citep{Phil03,Zeib09,BC2010}.

\subsubsection{Cooperation}

People are predisposed to ascribe a mind to artefacts exhibiting a certain degree of social connections and human likeness \citep{Waytz}. An artificial agent looking, moving, and behaving as a human may elicit cooperation, so the robot-society loop closes, which promises significant benefits \citep{REF15}. For instance, during the movement of a humanoid agent in a crowd, at the risk of a collision human pedestrian will cooperate with it by deviating from their initial trajectories. 

The cooperation among human pedestrians has been initially addressed by means of heuristic arguments. Models simulating physical interaction among people  successfully reproduced different phenomena of self-organi\-za\-tion observed in human crowds \citep{REF13, Hel2000, Dye2008}. Simple multi-agent interaction rules can explain complex behavior with emerging collective avoidance, formation of unidirectional lanes, and stop-and-go waves \citep{REF15}. Considering the human-robot interaction in low-density crowds it has been shown that potential fields created on the basis of proximity among pedestrians and rapid replanning can provide suitable trajectories \citep{REF14}. In dense but unstructured  environments \citet{REF17} have shown that cooperation is essential for feasible robot navigation. They postulated that humans tend to select optimal trajectories and provided an explicit statistical model for interaction among humans and robots \citep{REF18}. Empirical human trajectories have been also used by  \citet{REF16} for fitting a decision-making model.

In this work we apply the PfCA paradigm to cooperative navigation in social scenarios, showing that 1) the compact cognitive maps support recursive cognition, allowing mutual interaction between the agent and the environment, 2) the compact cognitive maps reflect the cooperation, changing the cognition acquired by the agent about the situation and the decisions to be made on it, and 3) most of the essential information about cooperation, as agent effort during navigation or safety of its decisions, is contained into the compact cognitive map, so no external (i.e. by means of motor actions) or internal (i.e.simulated) executions of the adopted strategies (trajectories) are required to evaluate the decisions. In addition, from a technical point of view, we propose an improved algorithm for obtaining the compact cognitive maps (both in cooperation and coexistence situations), making it faster and more general (see Apendix \ref{App1}).

\begin{figure*}
\begin{center}
\includegraphics[width= 0.9\textwidth]{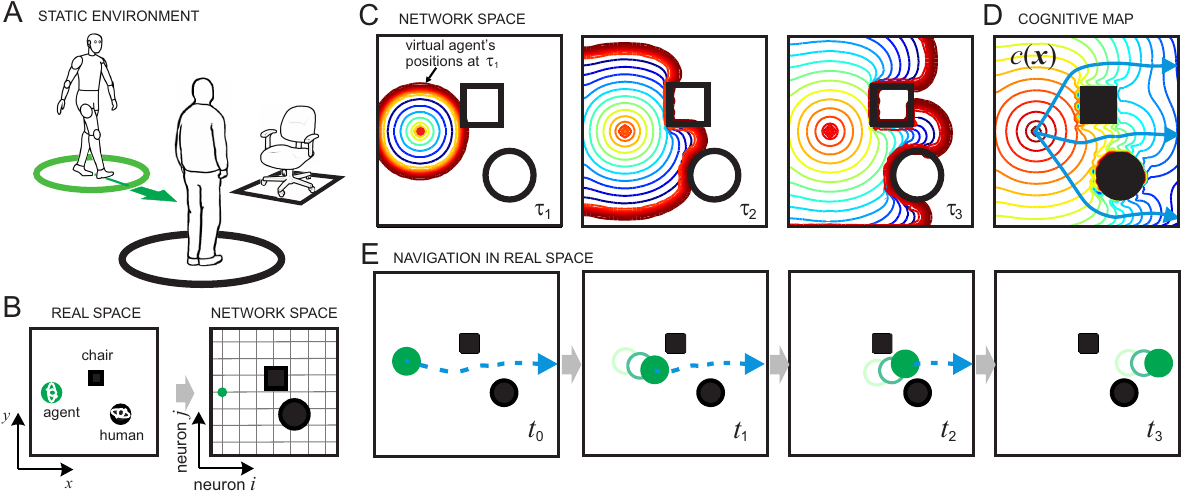}
\caption{Cognition through compact cognitive maps in a \emph{static environment}. A) A humanoid agent (green circle) walks avoiding collisions with a static human (black circle) and a chair (black square). B) The situation is mapped from the real space (left) to the network space (right) described by a 2D neural lattice. C) A wavefront propagating in the lattice simulates multiple agent's trajectories (three snapshots at mental time $\tau=\tau_{1,2,3}$). The front  explores the environment and creates a gradient profile. D) Final cognitive map with effective obstacles (in black). Going up the gradient the agent can reach the target avoiding obstacles (blue arrowed lines). E) Example of navigation. The agent follows one of the possible trajectories (superimposed frames with increasing green intensity correspond to progressively increasing time instants). 
\label{fig_CIR_Static}}
\end{center}
\end{figure*}

%%%%%%%%%%%%%%%%%%%%%%%%%%%%%%%%%%%%%%%%%%
%%%%%%%%%%%%%%%%%%%%%%%%%%%%%%%%%%%%%%%%%%
%%%%%%%%%%%%%%%%%%%%%%%%%%%%%%%%%%%%%%%%%%

\section{Cognition through compact cognitive maps}
\label{CCM}

Neural network implementation of cognition of static and dynamic situations may differ significantly.  Nevertheless, as we showed earlier  the concept of compact internal representation provides  an elegant way to unify  descriptions of both static environments and dynamic situations  \citep{BC2010}. In this Section we conceptually revisit the problem of how cognition may emerge in a bio-inspired neural network and introduce the notion of compact cognitive maps (for technical details see Appendix \ref{App1}). 

\subsection{Cognitive maps for static environments}
\label{SecCM}

Let us consider a situation sketched in Fig. \ref{fig_CIR_Static}A. A walking humanoid agent comes across two obstacles: a human and a chair.  The human and the chair stay immobile and therefore the agent is in a \emph{static environment}. Then the navigation can be fulfilled by using standard cognitive maps [see e.g.  \citep{StCogMaps}]. Earlier we provided a neural network implementation of this general concept \citep{BC2010}.

In real space the chair occupies some space whereas the agent and the human are represented by their personal areas \citep{PersonalArea}  [Fig. \ref{fig_CIR_Static}B, left panel]. To represent ``mentally" the real space we now introduce a 2D neural network, an $(n\times n)$-lattice of locally coupled neurons (Fig. \ref{fig_CIR_Static}B, right panel). This, so-called causal neural network (CNN, Appendix \ref{CNN_APP}), receives as an input the spatial configuration of the real space. In the network space 
\begin{equation}
\label{D}
D = \{(i,j) : \ \ i,j=1,2,\ldots n\}
\end{equation}
the agent is reduced to a single neuron, while its dimension is properly added to the obstacles' dimensions, thus proportionally increasing their sizes \citep{Lozano}. 

In the network space $D$ the agent has to create a cognitive map of the environment. It is fulfilled by virtual simulation of all possible agent's movements using a wave process. Since the agent can walk in any direction, its  virtual positions (locations occupied by virtual agents) at the next time step will form a circle with the agent in the center (Fig. \ref{fig_CIR_Static}C). The radius of this circle will grow with time as virtual agents will move away from the center. Thus, the process of mental exploration of the environment can be described by a solitary wave propagating in the network outward the agent's  initial position. The wavefront detects all obstacles, rounds them, and hence finds possible paths among them. We note that this is achieved efficiently, in one run independently on the complexity of the environment.

\begin{figure*}
\begin{center}
\includegraphics[width= 0.9\textwidth]{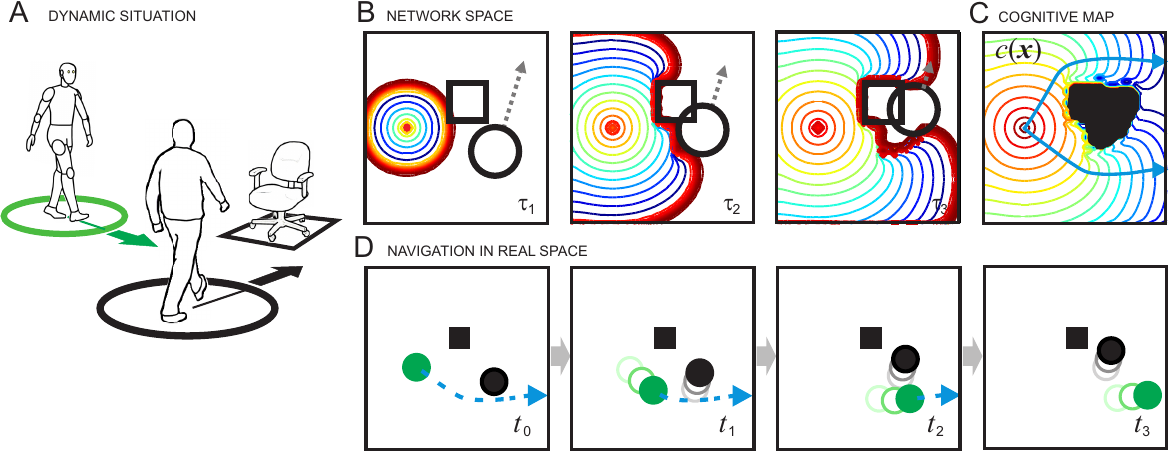}
\caption{Cognition through compact cognitive maps in a \emph{dynamic situation}. A) Same as in  Fig. \ref{fig_CIR_Static}A, but now the human walks towards the chair. B) Simulation of the agent's movements (wavefront) and matching them with obstacles' movements (human's trajectory, dashed line, is predicted by the TMNN). Collisions of the wavefront and virtual obstacles produce effective obstacles.  C) Compact cognitive map (the time dimension has been compacted) with static effective obstacle (black). Going up the gradient (blue arrowed curves) ensures collision-free walking. D) Agent navigating in the real space.
\label{fig_CIR_Dynamic}}
\end{center}
\end{figure*}

Each neuron $(i,j)$ records the time instants $c_{ij}$ when the wavefront  passes through it. Thus, we create a 2D potential profile
\begin{equation}
\label{C}
c: D \to \mathbb{R}
\end{equation}
or a cognitive map (Fig. \ref{fig_CIR_Static}D, contour curves). Going up the gradient $\nabla c$ (transversally from red to blue curves) the agent can follow one of the virtual trajectories (Fig. \ref{fig_CIR_Static}D, blue arrowed curves). These trajectories ensure colli\-sion-free walking in  the real space. Figure \ref{fig_CIR_Static}E illustrates one of the possible ways of navigation in this static environment.

%%%%%%%%%%%%%%%%%%%%%%%%%%%%%%%%%%%%%%%%%%%%
%%%%%%%%%%%%%%%%%%%%%%%%%%%%%%%%%%%%%%%%%%%%

\subsection{Compact cognitive maps for dynamic situations}
\label{SecCIR}

Figure \ref{fig_CIR_Dynamic}A sketches a situation  similar to that considered in Fig. \ref{fig_CIR_Static}A. However, now the human is going towards the chair and therefore the agent is in a \emph{dynamic situation}, which challenges standard cognitive maps.

The core of the Prediction-for-CompAction paradigm relies on two basic elements: i) Prediction of the movements of objects and ii) Simulation of all possible agent's trajectories. A special neuron network matches these processes and generates a  \emph{compact} cognitive map. Functionally this map is equivalent to a standard cognitive map, i.e. it is a static structure given by Eq. (\ref{C}), which allows for tracing collision-free trajectories.

\subsubsection{Prediction of object trajectory}
\label{TMNN}

To predict trajectories we use a dynamic memory  implemented in a so-called trajectory modeling neural network (TMNN, Appendix \ref{APPSecCIR}) \citep{BC2010,INTECH2011}.

A trajectory of a moving object (e.g. of the human in Fig. \ref{fig_CIR_Dynamic}A) is a function of time $s:\mathbb{R}\to\mathbb{R}^2$ that  can be approximated by a polynomial [similar to the spline method used by \citet{REF16}]:
\begin{equation}
\label{Taylor}
s(t) \approx s(0) + s'(0)t + \frac{s''(0)t^2}{2}
\end{equation}
Such an assumption is valid, at least, for inanimate objects and, as we will see below, for humans under the AvUs paradigm. The TMNN  predicts future object's locations, $\tilde{\vec{s}}$, by iterating a linear map $(W^k\vec{s}_0)_{k\in \mathbb{N}}$, where $W$ is a matrix describing couplings among neurons and $\vec{s}_0=(s(0), s'(0), s''(0))^T$ is the vector of initial momenta of the object (i.e. position, velocity, and acceleration). Then  
\begin{equation}
\label{PRED}
\tilde{\vec{s}} = \tilde{\vec{s}}(\tau; \vec{s}_0)
\end{equation}
is the trajectory in the network space $D$ and mental time $\tau = kh$, where $h$ is the time step and $k\in \mathbb{N}$.

For correct predictions the TMNN must be trained, i.e. the connectivity matrix $W$ must be  properly tuned \citep{BC2010}. Once the learning is finished, the TMNN is ready to predict the movement of objects solely based on their positions acquired by the sensory system at the present ($t = 0$) and two time instants in the past ($t=-2h$ and $t=-h$). Such predictions are quite robust against sensory noise  \citep{IEEE2013}.

\subsubsection{Simulation of agent's movements and matching them with predicted trajectories of objects}
\label{CCM_2}

As in the static case the CNN  simulates all possible movements of the agent by a wavefront (Fig. \ref{fig_CIR_Dynamic}B). However, since now the human moves in the environment, his trajectory, $\tilde{\vec{s}}(\tau)$, is predicted by the TMNN and expected future positions are fed to the CNN (compare Figs.  \ref{fig_CIR_Static}C \textit{vs} \ref{fig_CIR_Dynamic}B). Collisions of the wavefront and virtual objects in the network space correspond to possible collisions of the agent with objects in the real space.  In the CNN these locations delimit effective obstacles (Fig. \ref{fig_CIR_Dynamic}B; Appendix \ref{EFFOBST}).

Once the network space has been explored, the dynamic situation is represented as a static map (Fig. \ref{fig_CIR_Dynamic}C):
\begin{equation}
c = c(\vec{x}; \tilde{\vec{s}}), \ \ \ \vec{x}\in D
\end{equation} 
The  mobile (human) and immobile (chair) objects are replaced by the corresponding effective obstacles (joint black area). The gradient profile (contour curves from red to blue) contains a virtually infinite set of pathways that can  be followed by the agent (Fig. \ref{fig_CIR_Dynamic}C shows two representative examples):
\begin{equation}
\vec{d} = \vec{d}(\vec{x}; \nabla c)
\end{equation} 
Note that by simply avoiding static effective obstacles in $c(\vec{x})$ the agent avoids collisions with the human and the chair. Thus, the selected trajectory can be converted to motor actions and the agent can navigate in the real space following the corresponding path, $d(t)$, (Fig. \ref{fig_CIR_Dynamic}D).

Compact cognitive maps generalize the traditional concept of  cognitive maps extending it to time-changing situations. They compress spatiotemporal information about what and where may happen into static structures. The decision-making scheme can be represented as a unidirectional chain: 
\begin{equation}
\label{COGN1}
\begin{tikzpicture}[baseline=(current  bounding  box.center)]
\node[shape=circle,fill=green!20] (n1) at (0,0) {$s(t)$};
\node[shape=circle,fill=blue!10] (n2) at (1.5,0) {$\tilde{\vec{s}}(\tau)$};
\node[shape=circle,fill=blue!10] (n3) at (3,0)  {$c(\vec{x})$};
\node[shape=circle,fill=blue!10] (n4) at (4.5,0)  {$\vec{d}(\vec{x})$};
\node[shape=circle,fill=green!20] (n5) at (6,0)  {$d(t)$};

\draw[->] (n1) -- (n2);
\draw[->] (n2) -- (n3);
\draw[->] (n3) -- (n4);
\draw[->] (n4) -- (n5);
\end{tikzpicture}
\end{equation}

We note that at first glance the situation presented in Fig. \ref{fig_CIR_Dynamic}A may be considered delusively simple. Some geometric-based methods could provide solutions to this path planing problem \citep{REF15}. However, in the presence of several moving objects the problem becomes practically unsolvable for geometric algorithms or they would provide suboptimal solutions. Indeed, any trajectory deviation caused by avoidance of the first obstacle will induce changes in the configuration of possible collisions with the second obstacle, etc. Thus, the calculation diverges. Nevertheless, our neural network approach efficiently resolves navigation problems of practically arbitrary complexity.\footnote{Examples, simulations, and videos are available at http://www.mat.ucm.es/\%7Evmakarov/research.php}

%%%%%%%%%%%%%%%%%%%%%%%%%%%%%%%%%%%%%%%%%%
%%%%%%%%%%%%%%%%%%%%%%%%%%%%%%%%%%%%%%%%%%
%%%%%%%%%%%%%%%%%%%%%%%%%%%%%%%%%%%%%%%%%%

\section{Cognitive navigation in social environments}
\label{Sec3}
The agent's behavior in an environment with other cognitive agents (humans, animals, and/or robots) will depend on whether or not it can elicit cooperation of other parties. As we will see below there is a crucial difference between the AvUs and CoUs cognitions.

%%%%%%%%%%%%%%%%%%%%%%%%%%%%%%%%%%%%%%%%%%%%
\subsection{Machine Avoids Us (AvUs): Noncooperative navigation}

An AvUs agent expects no cooperation from humans and hence should move among people without disturbing them, ``making all the work'' to navigate safely. Thus, an AvUs  agent  can consider humans as ``inanimate" but moving objects, i.e. agent's  decisions/actions, $d(t)$, produce \emph{no feedback} to the human's trajectory, $s(t)$.  

\begin{figure}[htbp]
\begin{center}
\includegraphics[width= 0.47\textwidth]{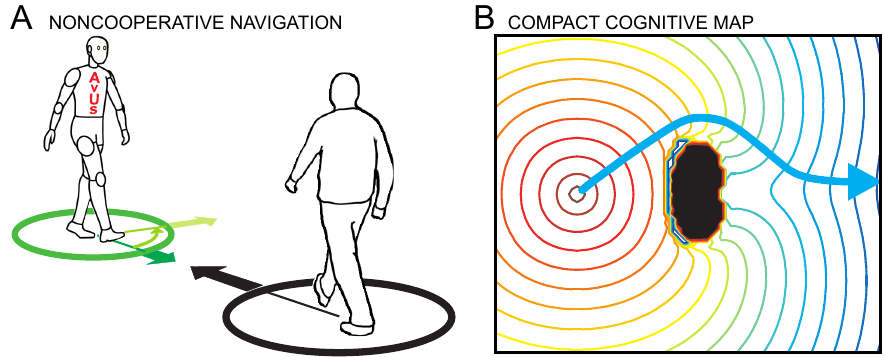}
\caption{Noncooperative navigation (AvUs paradigm). A) An agent expects no cooperation from a human. At the risk of a collision the agent steps away (light green arrow), while the human goes straightforward (black arrow). B) Compact cognitive map. The effective obstacle (black area) forces the agent to steer its trajectory (blue arrowed curve).
\label{Fig4a}}
\end{center}
\end{figure}

Figure \ref{Fig4a}A illustrates an AvUs agent approaching a human. The agent assumes noncooperative behavior of its human partner. Thus, the predicted human's movement depends uniquely on the initial conditions and will not be affected by the agent's decisions (given that the human will not suddenly change the trajectory). Thus, the basic Prediction-for-CompAction concept described in Sect. \ref{SecCIR} is sufficient to resolve this situation. 

Figure \ref{Fig4a}B shows the resulting compact cognitive map (the process of  creation is analogous to that shown in Fig. \ref{fig_CIR_Dynamic}B). The map contains a relatively big effective obstacle (black area) representing possible collisions with the human. Thus, the agent must steer around the effective obstacle to guarantee safety of the human and itself (Fig. \ref{Fig4a}B, blue arrowed curve).

%%%%%%%%%%%%%%%%%%%%%%%%%%%%%%%%%%%%%%%%%%%%
\subsection{Machine Cooperates with Us (CoUs): Cooperative navigation}

\begin{figure*}
\begin{center}
\includegraphics[width= 0.9\textwidth]{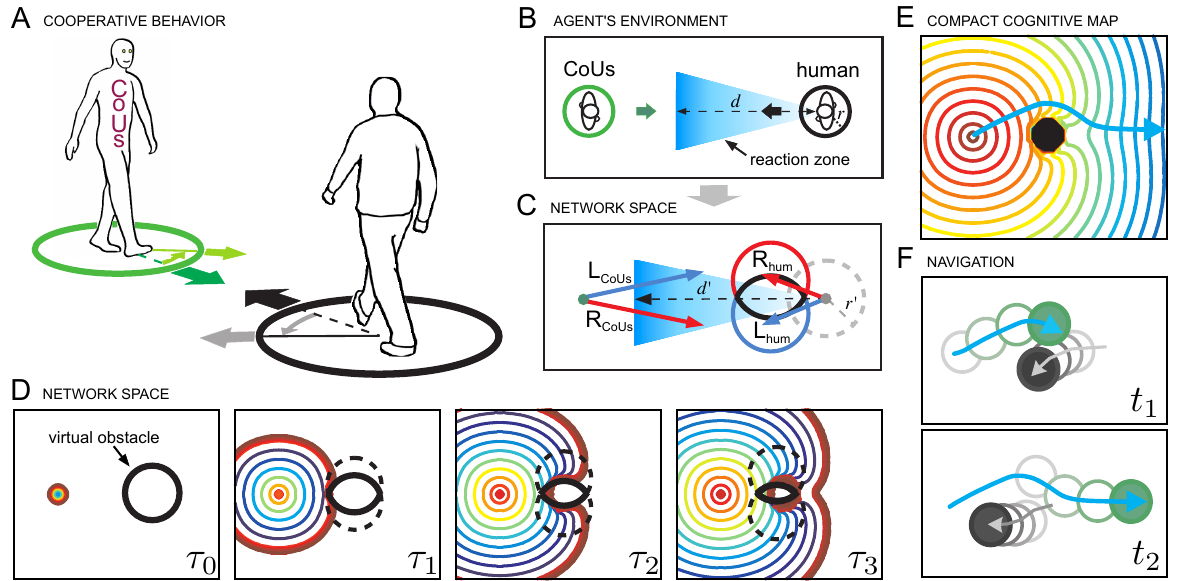}
\caption{Cooperative navigation (CoUs paradigm). A) Same as Fig. \ref{Fig4a}A, but now the human cooperates in collision avoidance (light grey arrow). B) Cooperation  occurs only if the CoUs agent enters the human's reaction zone under a proper angle. C) The agent can go either to the  right, $\rm R_{CoUs}$, or to the left, $\rm L_{CoUs}$, expecting the human response $\rm R_{hum}$ and $\rm L_{hum}$, respectively. The intersection of two personal zones forms a virtual obstacle to be avoided (area delimited by black solid curve). D) The process of mental exploration of possible movements (note decreasing size of the virtual obstacle). E)  Cooperation reduces the effective obstacle in the compact cognitive map (compare to Fig. \ref{Fig4a}B) and enables more efficient navigation. F) Example of navigation (superimposed frames with increasing color intensity correspond to progressively increasing time instants). 
\label{fig1}}
\end{center}
\end{figure*}

Let us now assume that humans are predisposed to cooperate with a properly looking and behaving humanoid agent. At the risk of a collision both the human and the agent will make a step out from their straight trajectories thus helping each other (Fig. \ref{fig1}A). 

The original Prediction-for-CompAction  paradigm assumes the feedforward information flow (\ref{COGN1}). However, now the agent's decision (i.e. the chosen trajectory $d(t)$) will have a feedback to the movement of the human, $s=s(t; d(t))$, which closes the information loop through the environment and the process becomes recursive: 
\begin{equation}
\label{COGN2}
\begin{tikzpicture}[baseline=(current  bounding  box.center)]
\node[shape=circle,fill=green!20] (n1) at (0,0) {$s(t)$};
\node[shape=circle,fill=blue!10] (n2) at (1.5,0) {$\tilde{\vec{s}}(\tau)$};
\node[shape=circle,fill=blue!10] (n3) at (3,0)  {$c(\vec{x})$};
\node[shape=circle,fill=blue!10] (n4) at (4.5,0)  {$\vec{d}(\vec{x})$};
\node[shape=circle,fill=green!20] (n5) at (6,0)  {$d(t)$};
\node[shape=rectangle,fill=red!20] (Env) at (3,-1)  {environment};

\draw[->] (n1) -- (n2);
\draw[->] (n2) -- (n3);
\draw[->] (n3) -- (n4);
\draw[->] (n4) -- (n5);
\draw[color=red!40,very thick] (n5) -- (6,-1) -- (Env) -- (0,-1);
\draw[->,color=red!40,very thick] (0,-1) -- (n1);
\end{tikzpicture}
\end{equation} 
This significantly challenges  cognition. Indeed, the agent should simulate all possible decisions (i.e. it should create $c(\vec{x})$ and then $\vec{d}(\vec{x})$ on the basis of $\tilde{\vec{s}}(\tau)$). In turn, each decision, $d(t)$, will provoke a reaction of the human, thus changing $s(t)$ and hence the predicted trajectory $\tilde{\vec{s}}(\tau)$. Then this new prediction should be taken into account to create new $c(\vec{x})$, etc. 

To cope with recursive cognition the agent must be able to model possible humans' actions using an internal loop:

\begin{equation}
\label{PRED3}
\begin{tikzpicture}[baseline=(current  bounding  box.center)]
\node[shape=circle,fill=green!20] (n1) at (0,0) {$s(t)$};
\node[shape=circle,fill=blue!10] (n2) at (1.5,0) {$\tilde{\vec{s}}(\tau)$};
\node[shape=circle,fill=blue!10] (n3) at (3,0)  {$c(\vec{x})$};
\node[shape=circle,fill=blue!10] (n4) at (4.5,0)  {$\vec{d}(\vec{x})$};
\node[shape=circle,fill=green!20] (n5) at (6,0)  {$d(t)$};
\node[shape=rectangle,fill=blue!20] (IM) at (3,-1)  {internal loop};
\node[shape=rectangle,fill=red!20] (Env) at (3,-2)  {environment};

\draw[->] (n1) -- (n2);
\draw[->] (n2) -- (n3);
\draw[->] (n3) -- (n4);
\draw[->] (n4) -- (n5);
\draw[color=blue!40,very thick] (n4) -- (4.5,-1) -- (IM) -- (1.5,-1);
\draw[->,color=blue!40,very thick] (1.5,-1) -- (n2);
\draw[color=red!40,very thick] (n5) -- (6,-2) -- (Env) -- (0,-2);
\draw[->,color=red!40,very thick] (0,-2) -- (n1);
\end{tikzpicture}
\end{equation}

\subsubsection{Heuristic model of human Perception-for-Action}
\label{HEURISTIC}
Earlier studies have shown that during walking humans perceive possible obstacles within certain visual angle and can estimate the time to collision \citep{Bib18,Bib19}. If a CoUs agent crosses the reaction zone (Fig. \ref{fig1}B) then the human tends to cooperate. To describe phenomenologically the human's behavior we adopt the following heuristics [similar to  \cite{Guy2011}]:

\begin{itemize} 
\item 
A human will cooperate in collision avoidance only if another agent  crosses his reaction zone under the angle $\varphi_{\rm crss} \in (-5^\circ,5^\circ)$ with the main visual axis (dashed line in Fig. \ref{fig1}B). Otherwise no cooperation is expected, i.e. if $|\varphi_{\rm crss}| \ge  5^\circ$ then the CoUs agent will be forced to behave like an AvUs one.

\item 
Under cooperative behavior, to avoid collision humans change their velocity vector: $\vec{v}_{\rm new} = \vec{v}_{\rm old} + \vec{w}$, where $\vec{v}_{\rm old}$ is the initial velocity and $\vec{w}$ is the normal vector to $\vec{v}_{\rm old}$ ($\|\vec{w}\| = \frac{1}{2}\|\vec{v}\|$, $\vec{w}\cdot \vec{v} = 0$). The direction of $\vec{w}$ depends on the  movement of the other agent.
\end{itemize}
We note that the second heuristic allows humans to keep the velocity in the target direction, which is useful, e.g., when moving in a group.

\subsubsection{Compact cognitive maps under CoUs paradigm}

In the network space the agent's dimension is reduced to a single neuron (Fig. \ref{fig1}C, green point), while the length of the human's reaction, $d$, and personal, $r$, zones increase, accordingly. In agreement with the Prediction-for-CompAction concept a wavefront propagating in the network space simulates all possible agent's decisions. The wavefront will cross the human's reaction zone both from the right and from the left of the midline. This means that the agent can avoid the human either from the right or from the left (Fig. \ref{fig1}C, trajectories $\rm R_{CoUs}$ or $\rm L_{CoUs}$, respectively). In the $\rm R_{CoUs}$ case the human is expected to turn right, $\rm R_{hum}$, whereas in the $\rm L_{CoUs}$ case he will go to the left, $\rm L_{hum}$. Since the human at the next time instant can occupy either of the two positions (Fig. \ref{fig1}C, red and blue circles), there appears uncertainty in the internal representation. The overlapping part of the corresponding personal zones (area delimited by black solid curve)  creates a virtual obstacle that the  agent must avoid. Note, that the remaining part of each personal zone is avoided due to the human-agent cooperation.

Figure \ref{fig1}D illustrates the process of generation of a compact cognitive map. A wavefront propagates outwards the agent's initial position. At the beginning ($\tau = \tau_0$, no interaction with the reaction zone) the human is represented as a moving circular obstacle, according to his personal zone. When the wavefront reaches the reaction zone, the human starts cooperating. Thus, the circular obstacle is split into two circles separating along the time course. This way the agent simulates the human responses adjusting $\tilde{\vec{s}}(\tau)$ according to $\vec{d}(\tau)$ [internal loop in  (\ref{PRED3})]. Then the virtual obstacle to be avoided is the intersection of these circles, which progressively diminishes (Fig. \ref{fig1}D, $\tau = \tau_{1,2,3}$). 

Figure \ref{fig1}E shows the final compact cognitive map. It contains a single effective obstacle (black area), which is significantly smaller than that in the case of the AvUs agent (Figs. \ref{fig1}E \textit{vs} \ref{Fig4a}B). Thus, the CoUs agent takes advantage of the human cooperation and gets more room for walking (blue arrowed curve).  Figure \ref{fig1}F shows an example of the CoUs navigation in the real space.

%%%%%%%%%%%%%%%%%%%%%%%%%%%%%%%%%%%%%%%%%%
%%%%%%%%%%%%%%%%%%%%%%%%%%%%%%%%%%%%%%%%%%
%%%%%%%%%%%%%%%%%%%%%%%%%%%%%%%%%%%%%%%%%%
\section{Navigation in crowd: Performance gain due to cooperation}

%%%%%%%%%%%%%%%%%%%%%%%%%%%%%%%%%%%%%%%%%%
\begin{figure}[thpb]
\centering
\includegraphics[width= 0.48\textwidth]{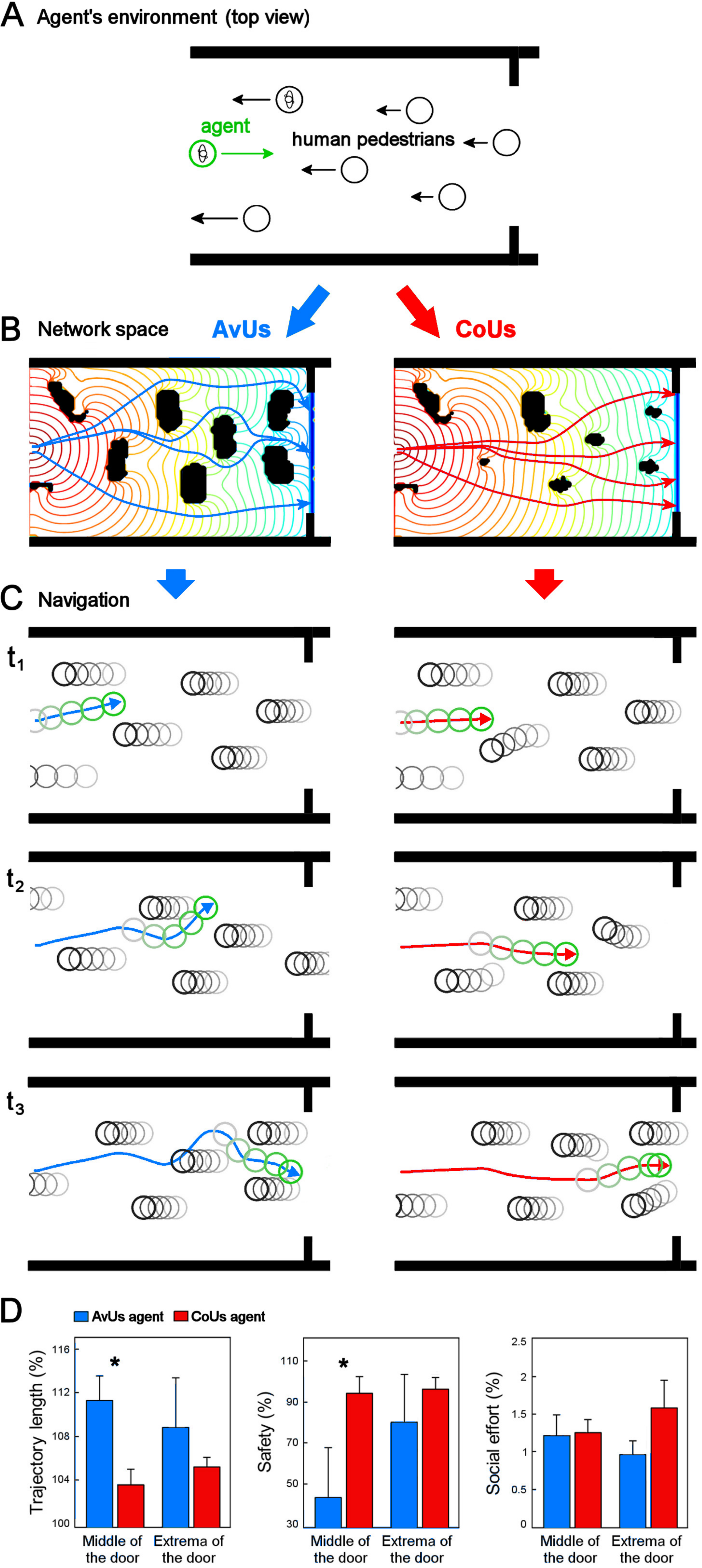}
\caption{Performance gain provided by cooperation in a cluttered crowd (CoUs \textit{vs} AvUs paradigm):  A) Initial situation. An agent (green circle) goes along a corridor to a door  against pedestrian flow (black circles). Arrows indicate the pedestrians' velocities. B) Compact cognitive maps created under non-cooperative (left) and cooperative (right) behaviors. Black areas correspond to effective obstacles. Arrowed curves show possible paths to the door. C) Examples of navigation of the AvUs (left) and CoUs (right) agents (superimposed frames with increasing color intensity correspond to progressively increasing time instants). D) Measures of the navigation performance (mean and std). Stars mark statistically significant difference, $p < 0.05$.}
\label{fig3}
\end{figure}

In order to study the performance gain and drawbacks provided by cooperation we simulated navigation of AvUs and CoUs agents in different human environments. In each situation we repeated internal simulations and decision making according to the two paradigms (Sect. \ref{Sec3}). Then we analyzed and compared the compact cognitive maps and agents' trajectories obtained from them.

\subsection{Benefits of CoUs strategy: Moving against cluttered pedestrian flow}

Let us model a situation frequently observed in the real world: an agent goes along a corridor against a pedestrian flow coming out of a door (Fig. \ref{fig3}A, circles correspond to personal areas). We simulated the pedestrian flow as a cluttered bunch of humans going in the same direction. The pedestrians' velocities increase as they move away from the door. Such behavior is natural when people have room to move \citep{REF13}. Besides, humans tend to optimize their trajectories and follow straight lines  \citep{REF17}, unless they cooperate for avoiding collisions (Sect. \ref{HEURISTIC}).

Figure \ref{fig3}B shows the compact cognitive maps and paths to the door found by the AvUs and CoUs agents. The two paradigms differ significantly in the representation of the dynamic situation. In general, effective obstacles in the AvUs map are bigger than in the CoUs one (Fig. \ref{fig3}B, left \emph{vs} right). Few big effective obstacles in the CoUs map appear due to violation of the condition for cooperation (Sect. \ref{HEURISTIC}). In this case the CoUs strategy is reduced to the AvUs one and hence we get similar effective obstacles in both maps. Thus, the CoUs agent gets more room for movement and, therefore, can plan more efficient trajectories to the door.

Figure \ref{fig3}C illustrates representative trajectories for both strategies. To avoid collisions the AvUs agent steers abruptly on its path to the door (left subplots). In the same situation the CoUs agent takes advantage of the human's cooperation. Its trajectory (right subplots) is significantly straighter. Note how pedestrians cooperate by deviating from straight lines and leaving more room to the CoUs agent.

To quantify the performance of the AvUs and CoUs strategies and to get deeper insight we separated agent's trajectories into two subsets: trajectories reaching the middle  part of the door (central door segment, $1/2$ width) and reaching the door's extrema (left and right door segments, $1/4$ width each).  Then we evaluated three performance measures (see Appendix \ref{App_B}).

\paragraph{Trajectory length.} 
Figure \ref{fig3}D (left) shows the mean normalized trajectory lengths for the AvUs and CoUs agents. As expected, the CoUs agent in general follows shorter paths to the door relatively to the AvUs. Moreover, the difference is statistically significant  if the agents target the door center [$p = 0.04$; here and below we used $t$-test \citep{ttest}], i.e. if they try to take the shortest way to the door, which favors cooperation and provides higher benefit. If the agents decide to go  around  the crowd to either of the door extrema the difference between the AvUs and CoUs strategies decreases ($p=0.254$). It occurs due to low  chances of cooperation with pedestrians on such a route. 

\begin{figure}[thpb]
\centering
\includegraphics[width= 0.48\textwidth]{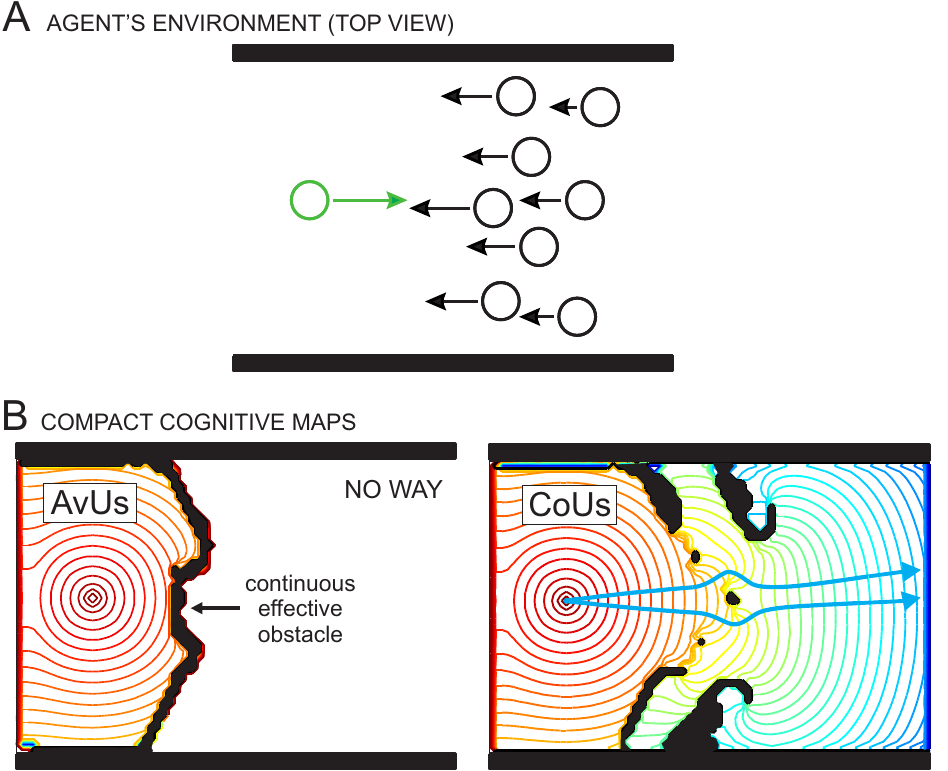}
\caption{Cooperation can help solve complex navigation problems. A) Initial situation. An agent (green circle) goes against a compact group of people. Arrows indicate pedestrians' velocities. B) Compact cognitive maps created by the AvUs (left) and CoUs (right) agents. In the noncooperative AvUs case an effective obstacle closes the corridor impeding navigation. In the cooperative CoUs case blue arrowed curves indicate two possible pathways.}
\label{FigColumna}
\end{figure}

\paragraph{Trajectory safety.} 
In many real situations the choice of a particular trajectory does not solely rely on its length, but also includes safety as an important factor. We defined the navigation safety as the fraction of the trajectory passing sufficiently far away from the effective obstacles, i.e. keeping the distance longer than some critical value. Figure \ref{fig3}D (middle) shows the  trajectory safety for the AvUs and CoUs agents. Surprisingly, cooperation provides better navigation safety, especially  for trajectories going to the door center ($p = 0.027$). We note that this  is not due to the trajectory straightness (length) but due to the reduced size of effective obstacles (Fig. \ref{fig3}B).

\paragraph{Mean social effort.} 
Until now we considered the performance gain of the CoUs strategy from the agent's viewpoint. However,  cooperation also implies elongation of trajectories of other pedestrians. Thus, as expected the CoUs strategy provides benefits to the agent at the cost of inconvenience for other pedestrians. Such a situation may be socially unstable. Thus, we measured the impact of cooperation on the ``society''  as the mean trajectory elongation averaged over all pedestrians (including the agent). In the case of targeting the door center the CoUs strategy produces no additional load to the social effort (Fig. \ref{fig3}D, right, $p = 0.9$). However, the social load increases (not significantly, $p=0.19$) for trajectories going to the door extrema.
Thus, in the case of targeting the door center (natural for human's behavior)  cooperation provides benefits in terms of the trajectory length (time and energy consumption) and safety with no additional load to the society effort.

%%%%%%%%%%%%%%%%%%%%%%%%%%%%%%%%%%%%%%%%%%%%%%%%

\subsection{CoUs strategy is necessary for navigation in dense crowd}

Let us now show that besides being convenient cooperation can be necessary for successful navigation. Figure \ref{FigColumna}A illustrates a  situation with an agent going against a compact group of pedestrians occupying the entire width of the corridor. The group leaves no room to an AvUs agent. Indeed, the compact cognitive map provides no solution (Fig. \ref{FigColumna}B, left), since the wavefront (simulating the possible agent's movements) can not ``leak" through pedestrians. This happens since under mental simulation the agent's personal area is added to the areas occupied by other pedestrians and holes among them disappear. The CoUs paradigm elicits cooperation. Humans move away and leave room to the agent, enough to steer among them  (Fig. \ref{FigColumna}B, right). Thus, cooperation may be not only beneficial but also necessary to reach the goal.

%%%%%%%%%%%%%%%%%%%%%%%%%%%%%%%%%%%%%%%%%%%%%%%%

\subsection{Abuse of cooperation may penalize CoUs strategy}

\begin{figure}[thpb]
\centering
\includegraphics[width= 0.48\textwidth]{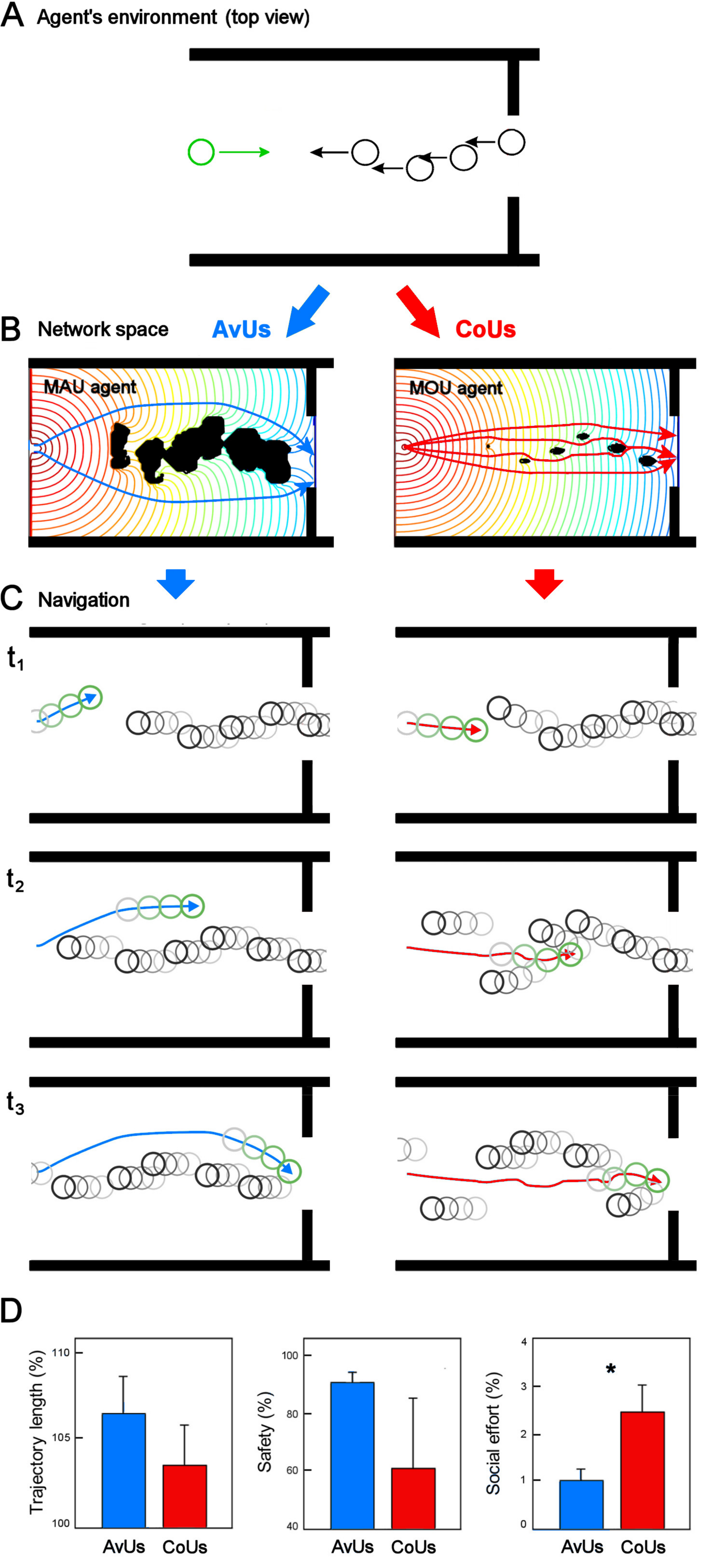}
\caption{Cooperative CoUs strategy can lose against non-cooperative AvUs navigation. A) Initial situation. An agent (green circle) goes along a corridor to a narrow door against a line-up pedestrian flow (black circles). Arrows indicate the pedestrians' velocities. B) Compact cognitive maps created by the AvUs (left) and CoUs (right) agents. Black areas correspond to effective obstacles. Arrowed curves show possible paths to the door. C) Examples of navigation of the AvUs (left) and CoUs (right) agents (superimposed frames with increasing color intensity correspond to progressively increasing time instants). D) Measures of the navigation performance (mean and std, star marks statistically significant difference).}
\label{FigFila}
\end{figure}

Figure \ref{FigFila}A shows a situation with a narrow door, which forces  pedestrians line up in a chain. Then the AvUs agent must go around the crowd since effective obstacles completely occupy the middle part of the cognitive map (Fig. \ref{FigFila}B, left). In the same situation the CoUs agent can go either  around or through the crowd. Cooperation leads to significantly smaller effective obstacles leaving holes among them (Fig. \ref{FigFila}B, right). Figure \ref{FigFila}C illustrates representative examples of trajectories performed by both agents. The CoUs agent ``pushes" pedestrians from its way and reaches the door following practically a straight line. 

Figure \ref{FigFila}D provides statistical properties of the AvUs and CoUs strategies. Since the door is narrow in this case we made no distinction between the center and extrema. The mean trajectory length is lower under cooperation, as expected (Fig. \ref{FigFila}D, left). However, this decrease is not significant ($p = 0.19$). The navigation safety decreases under cooperation, but again not significantly (Fig. \ref{FigFila}D, middle, $p=0.16$). This occurs because trajectories can go through the pedestrian chain and hence pass nearby humans, which increases chances of collision. Noteworthy, the cooperative strategy produces a statistically significant load to the social effort (Fig. \ref{FigFila}D, right, $p=0.02$), which is not acceptable for the robot behavior.

Thus, the cooperative CoUs strategy can lose against the noncooperative AvUs navigation. This paradoxical situation may occur in some particular spatial arrangements of the crowd. Indeed, the chain arrangement (Fig. \ref{FigFila}A) forces many pedestrians to stand away and the CoUs agent ``squeezes" through the crowd. Humans usually do not follow such sociopathic behavior, since it is socially not acceptable and increases the risk of collisions with many pedestrians. Therefore, abuse of cooperation by the CoUs agent may be inconvenient.

%%%%%%%%%%%%%%%%%%%%%%%%%%%%%%%%%%%%%%%%%%%%%%%%%%%%%%%%
%%%%%%%%%%%%%%%%%%%%%%%%%%%%%%%%%%%%%%%%%%%%%%%%%%%%%%%%
%%%%%%%%%%%%%%%%%%%%%%%%%%%%%%%%%%%%%%%%%%%%%%%%%%%%%%%%

\section{Discussion} 
\label{sec:conclusion}

Artificial cognition largely deals with the comprehension of relationships among elements in the environment. Nowadays the theoretical concept of cognitive maps, as a mean for understanding static situations, received strong experimental support  \citep{COGMAPS2013}. However, in dynamic situations spatial relationships among objects evolve in time. Therefore, the corresponding cognitive map should also change in time, which contradicts the very concept of a map. 

In this work first we discussed how the theory of compact internal representation \citep{BC2010} can generalize the concept of cognitive maps upon dynamic situations. The used neural architecture consists of two coupled neural networks. A recurrent neural network predicts positions of the obstacles for $t > 0$. These data are mapped into the other 2D neuronal lattice that simulates what will happen if the agent will take this or that trajectory. A wavefront propagating over the lattice collides with obstacles provided by the first network and forms a static potential field surrounding ``islands'' of effective obstacles (Fig. \ref{fig_CIR_Dynamic}). Thus, a dynamic situation can be ``mentally" represented as a static structure similar to a classical map. We called this process Prediction-for-CompAction (PfCA). The obtained compact cognitive map enables navigation avoiding collisions both with moving and static obstacles.   In this work we revised the lattice dynamics [earlier proposed by \cite{BC2010}] in such a way that the new neural network implementation makes no \emph{a priori} distinction between obstacles and targets. Thus, new compact cognitive maps are more universal, independent of the context. The same object in a map can be assigned as a target or as an obstacle and the agent can plan  a chasing or escaping actions.

The human behavior differs significantly from the behavior of inanimate but moving objects. Therefore, cognition of situations involving humans brings another dimension of complexity.  In this case the agent's actions depend on cognitive decisions made by humans and \emph{vice versa}. Thus, the cognitive process becomes recursive \citep{Recurs}.  In this article within the PfCA paradigm we addressed the first level of recursion. The agent assumes that its human partner under the risk of a collision will deliberately change his trajectory. Thus, the original concept of the trajectory prediction, based solely on initial conditions \citep{BC2010}, should be corrected for humans. Then we equipped  the agent with an internal heuristic model of the human behavior, similar to that proposed by \cite{Guy2011}. The model assumes that a human changes the velocity vector once the agent enters his reaction zone ($\vec{v}_{\rm new} = \vec{v}_{\rm old} + \vec{w}$). This prediction is introduced into the 2D neuronal lattice and, as in the inanimate case, a wavefront explores all possible agent's actions and generates a compact cognitive map describing recursive cognition (Fig. \ref{fig1}). Thus, the novel PfCA theory allows for coexistence of cognitive agents and cooperation emerges as a product of the neural network dynamics.

In order to investigate the plausibility and performance of the proposed theory we simulated navigation of an agent in different social environments. Two behavioral paradigms have been tested: ``machine avoids us" (AvUs) and ``machine cooperates with us" (CoUs). For navigation under either of the paradigms the agent builds a compact cognitive map. The map structure (position and size of effective obstacles) depends on the assumed human behavior, either cooperative or noncooperative. We have shown that the CoUs strategy sizably reduces effective obstacles comparing to the AvUs map (Fig. \ref{fig3}). Therefore, in the same situation a CoUs agent gets more room for movement than an AvUs one. Thus, in many realistic situations, including navigation in cluttered pedestrian flows, a CoUs agent can choose significantly shorter paths to the target (and hence spend less time and energy). Moreover, the navigation safety (i.e. chances to avoid accidental collisions) increases under cooperation. 

Cooperative navigation usually forces pedestrians turn aside from their straight courses. This inevitably elongates trajectories of all partners. The acceptable degree of cooperation in human society and the ensuing performance drop depend on many factors (e.g. cultural, emotional or environmental). Then the CoUs strategy will be socially acceptable if its cost to human pedestrians will be reasonable (similar to human cooperation). Thus, we measured the mean impact of the CoUs strategy on the society, i.e. the mean elongation of the trajectories averaged over all pedestrians including the agent. We have shown that in unstructured, low-dense pedestrian flows (such as in Fig. \ref{fig3}) the CoUs agent produces no additional load to the society effort. Thus a PfCA equipped artificial agent can behave as ``one of us". 

Another advantage of the CoUs strategy appears in extremely dense environments. It may happen that a group of people leaves no room for passing through (Fig. \ref{FigColumna}). Then the AvUs strategy provides no way to the target, whereas the cooperative behavior of people helps a CoUs agent find a path through the bunch of people.  It is worth noting that the PfCA allows  detecting situations with no solution, as e.g. in Fig. \ref{FigColumna} under AvUs. Thus, we can estimate chances of success without taking unnecessary risks of pursuing unreachable targets. This is a key attribute of the global decision-making,  distinguishing evolved animals from simple living beings \citep{IEEE2013}.

Though cooperation is usually beneficial, humans do not always exploit it. We  provided an example of such a situation: an agent going against a dense, chain-like pedestrian flow coming from a narrow door (Fig. \ref{FigFila}). We have shown that in this situation the CoUs strategy loses against the noncooperative AvUs navigation. This paradoxical result occurs due to the spatial arrangement of the crowd, which constrains many pedestrians to give pass to the CoUs agent squeezing its way through the crowd. Humans usually avoid such sociopathic decision, since it highly increases the risk of collisions with many pedestrians and is socially reproachable. Therefore, in certain situations cooperation may be inconvenient. Then, prior to taking any action, an advanced agent should  ``mentally" evaluate through cognitive maps the advantages and risks of the CoUs and AvUs behaviors and adopt correct decision.

In conclusion, the proposed neural network architecture provides cognitive skills necessary for  versatile and efficient navigation in social environments. The introduced AvUs and CoUs navigation strategies are complementary and their use depends on the context. An advanced artificial agent should test both of them and select the appropriate for each situation. Moreover, the PfCA theory enables high-level cognitive abilities like learning and memory, which is a sensible challenge for any artificial system \citep{IEEE2013}. Then successful experiences can be learnt and transformed into  efficient automatic-like behaviors. The proposed approach has been tested focusing on the agent's internal simulation and decision-making.  Its capacity in dealing with uncertainty and deviation of humans from the assumed  cooperative or noncooperative behavior is an open question left for further studies.

%%%%%%%%%%%%%%%%%%%%%%%%%%%%%%%%%%%%%%%%%%
%%%%%%%%%%%%%%%%%%%%%%%%%%%%%%%%%%%%%%%%%%
%%%%%%%%%%%%%%%%%%%%%%%%%%%%%%%%%%%%%%%%%%

%\begin{appendices}
\appendix
\section{Neural networks implementing compact cognitive maps}
\label{App1}
The main principles and details of the compact internal representation (CIR) have been discussed elsewhere \citep{BC2010,IEEE2013}. Briefly, the CIR is generated by a causal neural network (CNN) that receives as an input locations of all objects in the arena predicted by the trajectory modeling neural network (TMNN). The joint network dynamics forms effective static elements (e.g. effective obstacles) and a potential field $c$  in the network space $D$, which constitute a compact cognitive map. Then the map can be used to trace trajectories to a target (including moving). We note that any object or place in the environment can be assigned as a target.

\subsection{Trajectory modeling}
\label{APPSecCIR}

The TMNN implements a dynamic memory \citep{Mak08}. It  models object trajectories by quadratic polynomial [see Eq. (\ref{Taylor})]. Then in 2D space we have two components $s(t) = (x(t),y(t))$ modeled by two TMNNs. Each TMNN consists of three recurrently coupled neurons with external input $\xi(k) \in \mathbb{R}^3$ and output $\eta (k + 1) \in \mathbb{R}^3$, where $k=0,1,2,\ldots$ is the discrete \emph{mental} time \citep{BC2010}. 
The TMNN dynamics is given by
\begin{equation}
\label{eq1}
\eta(k+1) = \left\{
\begin{array}{ll}
\xi(k), \ \  & \mbox{if } \ | \xi(k) | > \delta \\
W\eta(k), \ \ &  \mbox{otherwise}  
\end{array}
\right.
\end{equation}
where $W \in \mathcal{M}_{3 \times 3}(\mathbb{R})$ is the coupling matrix and $\delta$ is the tolerance constant ($\delta = 10^{-6}$).
The TMNN operates in two phases: learning and prediction. Under learning, the TMNN receives at the input object trajectory  $\xi(k) = (x(k), v(k), a(k))^T$, where $x(k)$, $v(k)$, and $a(k)$ are the position, velocity, and acceleration of the object, respectively. Then the interneuronal couplings are updated according to: 
\begin{equation}
\label{LEARN}
\begin{array}{ll}
W(k+1) = & W(k) (I - \epsilon \xi(k - 1) \xi^T (k - 1)) \\
		&+ \epsilon \xi(k) \xi^T(k - 1)
\end{array}
\end{equation}
where $\epsilon > 0$ is the learning rate. Previously we have shown that the learning process (\ref{LEARN}) converges, given that $\epsilon$ is small enough \citep{BC2010}. 

Once the learning is deemed finished, the TMNN can predict trajectories. The object initial moments $\xi(0)=(x(0), v(0),$ $a(0))$ are sent to the TMNN for $k=0$ and then $\xi(k) = 0$ for $k>0$. On the output we get the predicted trajectory: $\eta(k) = W^k \xi(0)$.

\subsection{Causal neural network}
\label{CNN_APP}

The CNN is a 2D lattice of FitzHugh-Nagumo neurons ($80 \times 80$ cells in numerical experiments). The lattice dynamics is given by:
\begin{equation}
\label{EqAL}
\begin{array}{l}
\dot{r}_{ij}=  q_{ij} \left ( f(r_{ij}) - z_{ij}  
 + d\Delta r_{ij} \right ), \ \ \ \ (i,j) \in D \\
\dot{z}_{ij} = \varepsilon (r_{ij} - 7z_{ij} - 2) 
\end{array}
\end{equation}
where $r_{ij}$ and $z_{ij}$ are the membrane potential and recovering variable of the $(i,j)$-th neuron, respectively.  Dots represent derivatives with respect to the mental time $\tau = hk$, $\Delta$ is the discrete Laplacian, and $f(r)$ is a cubic like nonlinear function. The system (\ref{EqAL}) is considered with Neumann boundary conditions. In numerical experiments we used $d = 0.2$, $\varepsilon = 0.04$, and $f(r) = (-r^3+4r^2-2r-2)/7$. The function $q_{ij}(\tau)$ describes effective objects and will be discussed in Sect. \ref{EFFOBST}.

At the beginning all cells are at rest ($r_{ij}(0) = z_{ij}(0) = 0$) except one. The neuron $(i_{a},j_{a})$ corresponding to the agent's location has no dynamics $q_{i_{a}j_{a}} = 0$ and hence $r_{i_{a}j_{a}}(\tau > 0) = r_{i_{a}j_{a}}(0) = 5$ (we remind that in the network space $D$ the agent is reduced to a single cell). 

\subsection{Compact cognitive map and effective objects}
\label{EFFOBST}
The TMNN predicts movements of the obstacles and targets in the environment, while the CNN matches this information with the process of simulation of agent's movements. 
 
A wavefront propagating from the agent position is generated in the CNN (see Fig. \ref{fig_CIR_Dynamic}B). It switches cells to upstate. The time $\tau = c$ when the cell $(i,j)$ crosses a threshold ($r_{ij}(c) = r_{\rm th}$) is stored. Thus behind the wavefront we obtained a potential field $\{c_{ij}\}$ (see also Eq. (\ref{C})). 

Let $\mathcal{B}(k)$ be a set of cells $\{(i,j)\}\in D$ occupied by obstacles and targets at the mental time $k$. Then we define the following iterative process:
\begin{equation}
\label{SetEffObs}
\Omega(k) = \Omega (k-1) \cup \delta\Omega(k), \ \ \ k = 1,2,\ldots ; \ \ \Omega (0) = \emptyset
\end{equation}
where 
$$
\delta\Omega(k) = \{ (i,j)\in D  : \ \ r_{ij}(kh) \in [1, 2], \ \ (i,j) \in \mathcal{B}(k) \}
$$
The set $\Omega(k)$ describes effective objects (obstacles and targets) in the network space $D$. It is dynamically created as the wavefront  explores $D$. The set grows (i.e. $\delta\Omega(k) \ne \emptyset$)  if the wavefront touches an object at $\tau = kh$.  Then we define the function $q(\tau)$ in Eq. (\ref{EqAL}) as:
$$
q_{ij}(\tau) = \left \{
\begin{array}{l}
0, \ \ \ \mbox{if } (i,j) \in \Omega(k)\\
1, \ \ \ \mbox{otherwise}
\end{array}
\right .
$$
The cells in $\Omega(k)$ will exhibit no dynamics, i.e. the effective objects are \textit{static} and the wavefront slips around them (Fig. \ref{fig_CIR_Dynamic}B, panels $\tau_{2,3}$). 

Once the exploration of $D$ has been finished, the created CIR of the dynamic situation represents a compact cognitive map (Fig. \ref{fig_CIR_Dynamic}C). It contains spatial relationships (a potential field $c$) structured by static effective objects. These effective objects contain critical information about possible collisions of the agent and obstacles (to be avoided) or targets (to be pursued).

\subsection{Trajectory tracing}

To obtain a trajectory we use the gradient descent method. Since compact cognitive map does not distinguish between obstacles and targets, we should designate one (or several) of the effective objects in the map to be a target. Then we start from some point at the effective target and go down the gradient $\gamma_{k+1} = \gamma_k - \nabla c$. The obtained trajectory ends at the agent's location (the deepest part of the potential). We note that by construction the potential $c(i,j)$ has no local minima and hence a solution always exists (Fig. \ref{fig_CIR_Dynamic}C, arrowed curves).

%%%%%%%%%%%%%%%%%%%%%%%%%%%%%%%%%%%%%%%%%%%%%
%%%%%%%%%%%%%%%%%%%%%%%%%%%%%%%%%%%%%%%%%%%%%
%%%%%%%%%%%%%%%%%%%%%%%%%%%%%%%%%%%%%%%%%%%%%
%%%%%%%%%%%%%%%%%%%%%%%%%%%%%%%%%%%%%%%%%%%%%

\section{Measures of navigation performance}
\label{App_B}
\subsection{Trajectory length} 

Let us assume that an agent should move from a starting point $p_{\rm A}=(x_{\rm A},y_{\rm A})$ to a target point $p_{\rm T}=(x_{\rm T},y_{\rm T})$. On its route the agent follows  a trajectory given by vertices $\{p_i\}_{i=0}^N$, such that $p_0=p_{\rm A}$ and $\|p_N - p_{\rm T}\| < \epsilon$ ($\epsilon$ is the navigation tolerance and $\| \cdot \|$ is the Euclidean norm). 
Then we quantify the trajectory length relative to the shortest straight path to the target:
\begin{equation}
\label{TR_LNGTH}
L =  \frac{1}{\|p_{\rm A} - p_{\rm T} \|}
\sum_{i = 1}^{N}\| p_{i} - p_{i-1}\|
\end{equation}
The closer the normalized length $L$ to $1$, the shorter the trajectory and the higher the navigation performance (i.e., faster arrival and smaller energy consumption).

\subsection{Trajectory safety} 
Let $d_{\rm crt}$ be a critical distance between the agent and obstacles at which the navigation is considered safe. Then we define the measure of safety of a trajectory $\Gamma$ as the ratio: 
\begin{equation}
\label{TR_SFT}
S = 1-\frac{{\rm card}(\delta \Gamma)}{{\rm card}(\Gamma)}
\end{equation}
where 
$$
\delta \Gamma = \{ p=(x,y) \in \Gamma : \ \ D(p, \Omega) < d_{\rm crt}\}
$$
where $\Omega$ is the final set of effective obstacles, defined by (\ref{SetEffObs}), and
$$
D(p, \Omega) = \inf_{\tilde{p} \in \Omega } \| p - \tilde{p}\| 
$$

\subsection{Mean social effort} 

We define the social effort as the average elongation of the normalized trajectories  of all pedestrians, including the agent:
\begin{equation}
\label{SCL_EFF}
E = \frac{1}{M+1} \sum_{i=0}^{M} (L_i - 1)
\end{equation}
where $M$ is the number of pedestrians and $L_i$ is the trajectory length (\ref{TR_LNGTH}) of  the $i$-th pedestrian ($i=0$ corresponds to the agent).  The higher the value of $E$, the greater the effort of the society members to navigate to their goals. 

%\end{appendices}

%\bibliographystyle{spbasic}
%\bibliography{refer_Cogn_Soc_Navig}

\end{document}